\documentclass[lettersize,journal]{IEEEtran}
\usepackage{amsmath,amsfonts}
\usepackage{algorithmic}
\usepackage{algorithm}
\usepackage{array}
\usepackage[caption=false,font=normalsize,labelfont=sf,textfont=sf]{subfig}
\usepackage{textcomp}
\usepackage{stfloats}
\usepackage{url}
\usepackage{verbatim}
\usepackage{graphicx}
\usepackage{cite}
\usepackage{booktabs}
\usepackage{tabularx}
\begin{document}

\title{LIR-LIVO: A Lightweight,Robust LiDAR/Vision/Inertial Odometry with Illumination-Resilient Deep Features}

\author{
Shujie Zhou, Zihao Wang, Xinye Dai, Weiwei Song and Shengfeng G$\text{u}^*$
}



\maketitle

\begin{abstract}
In this paper, we propose LIR-LIVO, a lightweight and robust LiDAR-inertial-visual odometry system designed for challenging illumination and degraded environments. The proposed method leverages deep learning-based illumination-resilient features and LiDAR-Inertial-Visual Odometry (LIVO). By incorporating advanced techniques such as 
uniform depth distribution of features enabled by depth association with LiDAR point clouds and adaptive feature matching utilizing Superpoint and LightGlue, LIR-LIVO achieves state-of-the-art (SOTA) accuracy and robustness with low computational cost. Experiments are conducted on benchmark datasets, including NTU-VIRAL, Hilti'22, and R3LIVE-Dataset. The corresponding results demonstrate that our proposed method outperforms other SOTA methods on both standard and challenging datasets. Particularly, the proposed method demonstrates robust pose estimation under poor ambient lighting conditions in the Hilti'22 dataset. The code of this work is publicly accessible on GitHub\footnote{https://github.com/IF-A-CAT/LIR-LIVO} to facilitate advancements in the robotics community.
\end{abstract}


\section{Introduction}
\IEEEPARstart{W}{ith} the rapid development of simultaneously localization and mapping (SLAM) technique and the advancement of various perception sensors, real-time navigation and high-precision mapping for platforms such as robots and unmanned aerial vehicles (UAVs) have become increasingly achievable\cite{Unified}. Among the commonly used SLAM sensors, LiDAR excels in directly measuring point distances and 3D coordinates, making it ideal for real-time 3D reconstruction and high-precision odometry through point cloud registration\cite{LOAM}. However, its lack of color information can be addressed by RGB cameras, which enrich semantic understanding and enable visual odometry using the Perspective-n-Point (PnP)\cite{vinsmono}. IMUs complement these sensors by providing high-frequency, precise pose estimates, aiding in camera keyframe initialization\cite{vinsmono} and correcting LiDAR keyframe distortions caused by motion\cite{fastlio2}. Real-world localization and mapping tasks frequently occur in environments with structural degradation or diminished visual features, where single-sensor solutions frequently fail to deliver reliable performance. To tackle these challenges, multi-sensor fusion strategies\cite{fastlivo,fastlivo2,R3live,lvisam,levins,Unified,srlivo,camvox} have garnered growing attention for leveraging the strengths of diverse sensors to mitigate accuracy degradation in challenging environments.

In the most advanced and state-of-the-art SLAM solutions, two common approaches stand out: the combination of inertial sensors with vision\cite{vinsmono}, and the combination of inertial sensors with LiDAR\cite{fastlio2,liosam,LOAM}. Recently, a more comprehensive solution integrating LiDAR, vision, and inertial sensors has gained attention as an effective and robust approach for achieving real-time, high-precision SLAM in complex and degraded environment\cite{fastlivo,fastlivo2,lvisam}. LIO-SAM\cite{liosam}, LOAM\cite{LOAM}, LEGO-LOAM\cite{Legoloam} utilize feature extraction methods to reduce the number of points in the LiDAR point cloud, making it feasible for efficient and accurate state estimation. However, feature extraction is also a relatively time-consuming process. Some LiDAR odometry adopt direct methods, such as FAST-LIO2\cite{fastlio2} and FAST-LIVO\cite{fastlivo}, which optimize point-to-plane errors to directly process raw point clouds. By bypassing the feature extraction step, these approaches achieve high efficiency and maintain accurate state estimation, making them well-suited for real-time applications. 

In SLAM tasks under LiDAR-degraded environments, LiDAR-inertial-visual fusion systems often struggle to maintain globally consistent and high-precision pose estimation due to limited robustness and excessive reliance on single-modality features. However, with the assistance of vision, the LiDAR-inertial-vision odometry (LIVO) system effectively address challenges encountered in LiDAR-degraded scenarios. Existing LiDAR-inertial-visual framework usually include two point maps, one is for LiDAR-inertial odometry subsystem, anothor is for visual feature points. The maintenance of two submaps is often time- and memory-intensive, making real-time operation difficult on resource-constrained devices. The visual component used in LiDAR-inertial-visual SLAM systems is typically adapted from traditional visual or visual-inertial odometry, which struggles with feature tracking under significant lighting variations.
\begin{figure}[h]
    \centering
    \includegraphics[width=\linewidth]{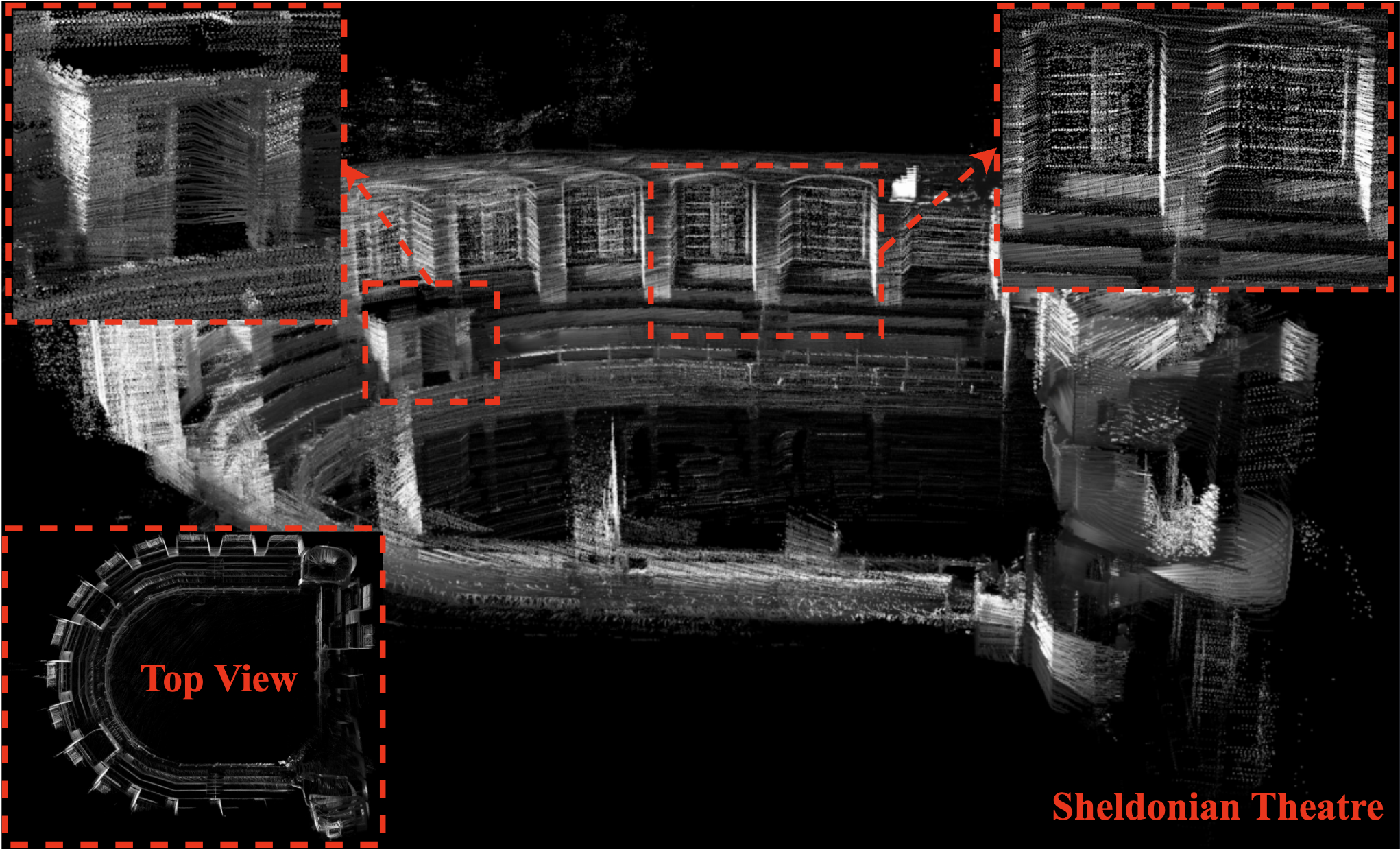}
    \caption{The 3D point cloud mapping results of LIR-LIVO on the Hilti'22 sequence ``Exp16 Attic to Upper Gallery 2''. The sequence was collected in the indoor environment of the Sheldonian Theatre in Oxford, categorized as a ``hard'' difficulty level.}
    \label{map}
\end{figure}

In this paper, we propose a lightweight, robust, illumination-resilient LiDAR-inertial-visual odometry system (LIR-LIVO), leveraging deep learning-based illumination-Resilient feature points and LiDAR-visual depth association, achieving low computational cost and high robustness. Our proposed method inherits the direct method of FAST-LIO2 for LiDAR-inertial odometry, achieving state estimation by optimizing point-to-plane distances. To more effectively leverage LiDAR point clouds for assisting the depth of visual feature points, we employ scan recombination to segment the high-frequency, sequentially sampled LiDAR point clouds into distinct LiDAR scans corresponding to camera sampling timestamps like \cite{srlivo,fastlivo2}, ensuring frequency consistency between LiDAR and the camera. Subsequently, we apply a sequential update algorithm to jointly update the state using the LIO and VIO systems. By associating LiDAR points with visual feature points, accurate 3D positions of visual features in space can be directly obtained, eliminating the need for visual triangulation and the update and maintenance of visual feature 3D points. This approach enhances both the efficiency of the VIO system and the accuracy of visual 3D points. Furthermore, a key innovation of the proposed algorithm lies in the integration of deep visual features extracted using convolutional neural networks and the utilization of the lightweight, efficient, and deep learning-based feature matching algorithm. This approach demonstrates superior performance in scenarios characterized by significant illumination variations and rapid motion, thereby significantly enhancing the robustness of LiDAR-inertial-visual odometry under challenging conditions such as bot LiDAR and illumination degraded environments. The main contributions of our work are listed as follows:
\begin{enumerate}
    \item A robust, lightweight, illumination-resilient LiDAR-inertial-visual odometry system is presented for tasks in LiDAR and vision challenging environments. LiDAR, visual and inertial sensors are tightly fused at the measurement level using an iterative Kalman filter, where the LiDAR scan point clouds provide precise depth information for visual features.
    \item The system leverages the deep learning-based SuperPoint algorithm for adaptive visual feature extraction and the LightGlue algorithm for efficient feature matching. Their exceptional robustness under significant illumination variations greatly enhances the resilience of the VIO subsystem, ensuring reliable performance in vision-degraded environments.
    \item The VIO subsystem, designed as a lightweight component with the assistance of LiDAR depth association, eliminates the necessity for maintaining and updating visual 3D landmarks and submaps. It maintains only a limited number of historical keyframes within a sliding window to construct reprojection errors for optimization. An optimized depth distribution of feature points is also applied to enhance pose estimation.
    \item The implementation of the proposed system has been open-sourced on GitHub to encourage community engagement and advance research in the related field.
\end{enumerate}

Real-time dense point cloud reconstruction conducted on the Hilti'22 ``Exp16 Attic to Upper Gallery 2'' is shown in Fig. \ref{map}. Since the sequence employs grayscale cameras, the generated point cloud maps are also grayscale. The remainder of this article is organized as follows: Section II reviews relevant literatures with respect to LiDAR-inertial-visual odometry. The overview of our proposed system is presented in Section III. The methods of LIR-LIVO are described in Section IV. Section V discusses the experiments and results, followed by the final conclusions in Section VI.

\section{Related Work}
In recent years, numerous LiDAR-inertial-visual fusion frameworks have been proposed to enhance the accuracy and robustness of SLAM systems under challenging conditions. \cite{stereoLiDAR} introduced a cascaded approach combining tightly coupled stereo VIO, LiDAR odometry, and a LiDAR-based loop-closing module to improve system performance. Lic-Fusion fused IMU data, sparse visual features, and LiDAR features within the multi-state constrained Kalman filter (MSCKF) framework, achieving online spatial and temporal calibration. LIC-Fusion 2.0\cite{licfusion2} further improved LiDAR point registration accuracy by introducing a plane-feature tracking algorithm across multiple LiDAR sweeps in a sliding window and refining sweep poses within the window. Graph-based optimization has also been explored in systems like LVI-SAM\cite{lvisam}, which tightly couples data from cameras, LiDAR, and IMUs. LVI-SAM allows independent operation of the vision and LiDAR modules when one fails or joint operation when both provide sufficient features. FAST-LIVO\cite{fastlivo} streamlined the fusion process by integrating LiDAR, camera, and IMU measurements into a single error-state iterated Kalman filter (ESIKF), enabling updates from both LiDAR and visual observations. FAST-LIVO2\cite{fastlivo2}, an upgraded version of FAST-LIVO, achieves higher pose estimation and mapping precision with exposure time estimation, reference patch update and normal refinement in real-world experiments. R3LIVE\cite{R3live}, built on R2LIVE\cite{R2live}, omitted the graph-based optimization module and introduced a color rendering module for dense color map reconstruction. SR-LIVO \cite{srlivo} utilizes sweep reconstruction to align LiDAR scans with camera timestamps, facilitating precise state estimation at imaging instances and significantly improving pose accuracy and computational efficiency. Likewise, FAST-LIVO2 synchronizes LiDAR point cloud frames to match the camera frame rate, ensuring temporally consistent updates between LiDAR and visual measurements, which enhances the cohesiveness of data fusion and overall system performance. 

In the field of VIO, many deep learning-based approaches have been employed to enhance robustness, accuracy, and performance. Droid-SLAM\cite{Droidslam}, for instance, has demonstrated an end-to-end visual SLAM system. However, its training process is time-consuming and computationally expensive. Beyond end-to-end methods, a growing body of research focuses on leveraging deep learning techniques to optimize the performance of the front-end, while relying on traditional filtering or factor graph optimization algorithms for the back-end. SupSLAM\cite{supslam}, a robust visual-inertial SLAM system that leverages SuperPoint\cite{superpoint} for feature detection and description, enabling accurate localization and mapping in challenging environments. AirVO\cite{airslam} utilizes SuperPoint for feature detection and SuperGlue\cite{superglue} for feature matching, achieving high robust front-end. In the new version of AirVO, known as AirSLAM, the LightGlue\cite{lightglue} algorithm is utilized as a replacement for SuperGlue, moderately improving the efficiency of feature matching. An increasing number of deep learning-based front-end feature extraction and tracking methods have been proposed, such as XFeat\cite{Xfeat}. With graphics processing unit (GPU) support, deep learning-based front-end systems have achieved a level of real-time performance comparable to traditional front-end approaches\cite{airslam,Xfeat}.
\section{system overview}
\begin{figure}[h]
    \centering
    \includegraphics[width=\linewidth]{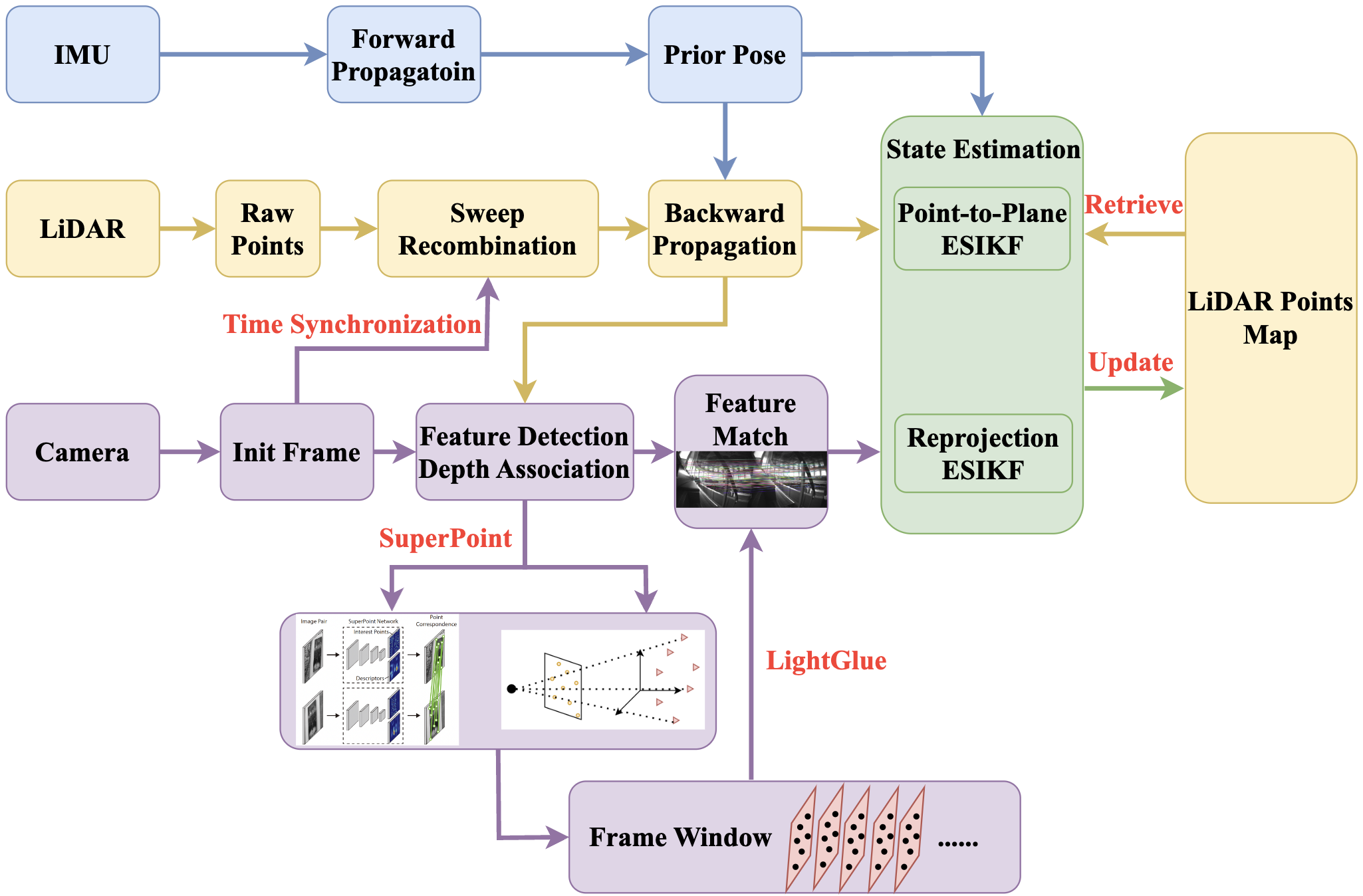}
    \caption{The framework of our LIR-LIVO. The LiDAR frame timestamps are synchronized with camera frame timestamps by sweeping recombination, enabling sequential updates of LiDAR and visual measurements. The visual frontend incorporates SuperPoint and SuperGlue frameworks, with optimized depth distribution of feature points. The black points in the frame window are the selected features with depth associated.}
    \label{framework}
\end{figure}

\begin{table}[h]
    \centering
    \caption{Some important Notations}
    \begin{tabularx}{0.5\textwidth}{ll}
    \toprule
        Notation & Explanation \\ \midrule
        $(\cdot)^l$ & The vector $(\cdot)$ in LiDAR coordinate.\\
        $(\cdot)^c$ & The vector $(\cdot)$ in camera coordinate.\\
        $(\cdot)^i$ & The vector $(\cdot)$ in IMU coordinate.\\
        $(\cdot)^g$ & The vector $(\cdot)$ in global coordinate.\\
        $T_l^c (R^c_l,t^c_l)$ & The extrinsics between LiDAR and camera coordinate.\\
        $T_l^i (R^i_l,t^i_l)$ & The extrinsics  between LiDAR and IMU coordinate.\\
        $T_c^i (R^i_c,t^i_c)$ & The extrinsics  between camera and IMU coordinate.\\
        \bottomrule
    \end{tabularx}
    \label{notations}
\end{table}
The important notations in the article are listed in Table \ref{notations}. The IMU coordinate is consistent with the body coordinate. Fig. \ref{framework}
demonstrates the overview of our system, which is composed of two main components: a direct method-based LiDAR module with the time synchronization of camera frame\cite{fastlio2}, and a lightweight visual module characterized by a deep learning frontend. The Sweep recombination is utilized to synchronize the timestamps of LiDAR frames and camera frames\cite{srlivo,fastlivo2}. Therefore, the ESIKF process is executed through sequential updates, enabling the LiDAR-updated pose to serve as a high-precision prior for visual processing. SuperPoint and LightGlue are used to construct the visual frontend, with feature point depths directly associated with LiDAR point clouds. During this process, the depths of feature points are filtered to ensure a uniform depth distribution. For LiDAR point cloud management, we adopt the same strategy as FAST-LIO2, utilizing ikd-Tree\cite{fastlio2}. Instead of directly maintaining a 3D landmark map for vision, we store the feature points of each frame along with their corresponding depths in the data structure of the respective camera frame, while maintaining a fixed number of camera frames using a sliding window. The point-to-plane error and reprojection error are used for pose estimation in LiDAR and vision, respectively.

\section{System Methods}
\subsection{Sweep Recombination}
To achieve precise synchronization between LiDAR and camera data in multi-sensor fusion systems, we apply a sweep recombination method\cite{srlivo,fastlivo2}. This technique ensures that both LiDAR raw points and camera images are aligned to the same frequency, enabling sequential and consistent state updates. Specifically, the end timestamp of the reconstructed LiDAR sweep is aligned with the timestamp of the captured image, which is critical for maintaining temporal consistency between the two modalities. The sweep recombination process is illustrated in Fig. 1. By disassembling the raw LiDAR sweep into individual point clouds and reconstructing it to align with the camera timestamps, we achieve complete synchronization between the LiDAR frames and camera frames. This enables seamless sequential updates for state estimation using both LiDAR and visual data at the same time in subsequent processing.
\begin{figure}
    \centering
    \includegraphics[width=\linewidth]{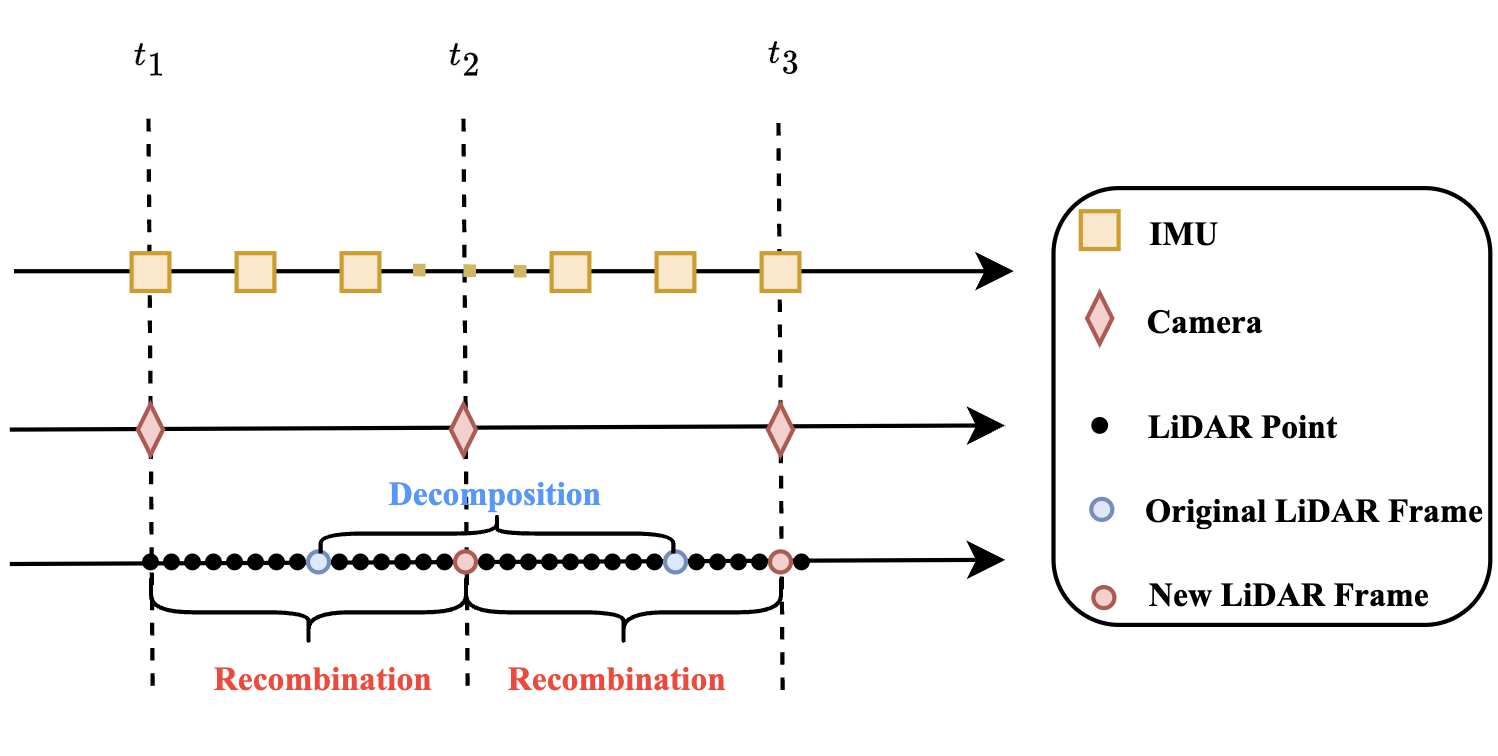}
    \caption{The process of sweep recombination. The original LiDAR frame is decomposed to construct a new one synchronized with camera frame.}
    \label{sweeprecombination}
\end{figure}
\subsection{Depth Association and Distribution}
Similar to the approaches described in \cite{high robust low drift,levins,depth enhance}, we employ a method to associate LiDAR points with depth information for visual features. A silding window is applied to maintain a certain number of LiDAR points within the camera field of view. We first project the LiDAR points to a unit sphere centered on the camera. A three-dimensional K-D tree is employed to store these points and search closest points with respect to the feature point, which is projected from image plane to the unit sphere. Once the five closest points are found, a further validity check is performed with a point-to-plane residual threshold. According to \cite{levins}, the normal vector of the plane in the unit sphere computed by five associated points can be expressed as 
\begin{equation}\label{plane}
\begin{cases}
    Ax =& b,\\
    A=&[p_1^c,p_2^c,p_3^c,p_4^c,p_5^c]^T,\\
    b=& [0,0,0,0,0],\\
    x =& (A^TA)^{-1} A^T b,
\end{cases}
\end{equation}
in which $x$ is consist of the normal vector $n$ and the plane offset $l$; $A$ and $b$ denote the coefficient matrix and prior residual, respectively; $p_i^c(i=1,2...5)$ are LiDAR points in the u. The problem above can be solved by the least square estimation. For each point used to calculate the plane, we check the point-to-plane residual with a threshold of 0.05m to prevent the occurrence of false associations, which is more strict than the proposed threshold mentioned in \cite{levins}. 

After validating the association of the five selected points $p_i^c$ and computing the normal vector $n$, one of five points is chosen to construct an equation with the query point for solving the depth $d$. The specific process is presented as follows
\begin{equation}
    \begin{cases}
        (p_i^c - p_0^c )^T \cdot n = (p_i^c - p_0^{u}\cdot d)^T\cdot n =0 ,\\
        d = ({p_i^c}^T \cdot n)/({p_0^u}^T\cdot n),
    \end{cases}
\end{equation}
where $p_0^c ,p_0^u$ denote the query 3D landmark of the corresponding feature point in the camera coordinate and the projection of it in the unit sphere, respectively. Consequently, the depth $d$ is associated with the corresponding visual feature.

To improve the accuracy and robustness of the visual odometry, we implemented a strategy that ensures an even distribution of feature points across multiple depth levels. Feature points at varying depths exhibit different sensitivities in pose estimation. Specifically, points at greater depths are more responsive to rotational pose estimation but have minimal impact on translational accuracy. However, points at shallower depths significantly enhance translational state estimation accuracy but are less effective for rotational estimation. Given two consecutive frames $j$ and $k$, the projection of a feature point $p^{p_j}$ in pixel plane from frame $j$ onto frame $k$ is mathematically described as
\begin{equation}
    \begin{cases}
    p^{p_{k}}&=\pi_{c}(\delta_k\left[R_j^k\frac{1}{\delta_j}\pi_{c}^{-1}(p^{p_j})+p_j^k\right])\\
    \frac{\partial p^{p_k}}{\partial p_{j}^{k}} &= {\pi_c}^{'} \cdot  \delta_k\\
    \frac{\partial p^{p_k}}{\partial \phi_{j}^{k}} &=  - {\pi_c}^{'} \cdot  \delta_k \left[
R_j^k \frac{1}{\delta_j}\pi_c^{-1}(p^{p_j}) \times \right]    \end{cases}
\end{equation}
where $p_j^k,R_j^k$ denote the translation and rotation from camera frame $j$ to camera frame $k$,respectively; $\delta_j ,\delta_k$ are inverse depths of the feature point in camera frame $j$ and $k$; $\pi_c$ and $\pi_c^{'}$ are the camera projection function and the corresponding first order derivative; $p^{p_j},p^{p_k}$ denote the feature points in pixel plane $j$ and $k$, respectively; $\phi_j^k$ is used to substitute the derivation of $R_j^k$, which serves as the perturbation to the rotation matrix $R_j^k$. The symbol $(\cdot)\times$ means the skewsymmetric matrix of $(\cdot)$. 

\begin{figure}
    \centering
    \includegraphics[width=\linewidth]{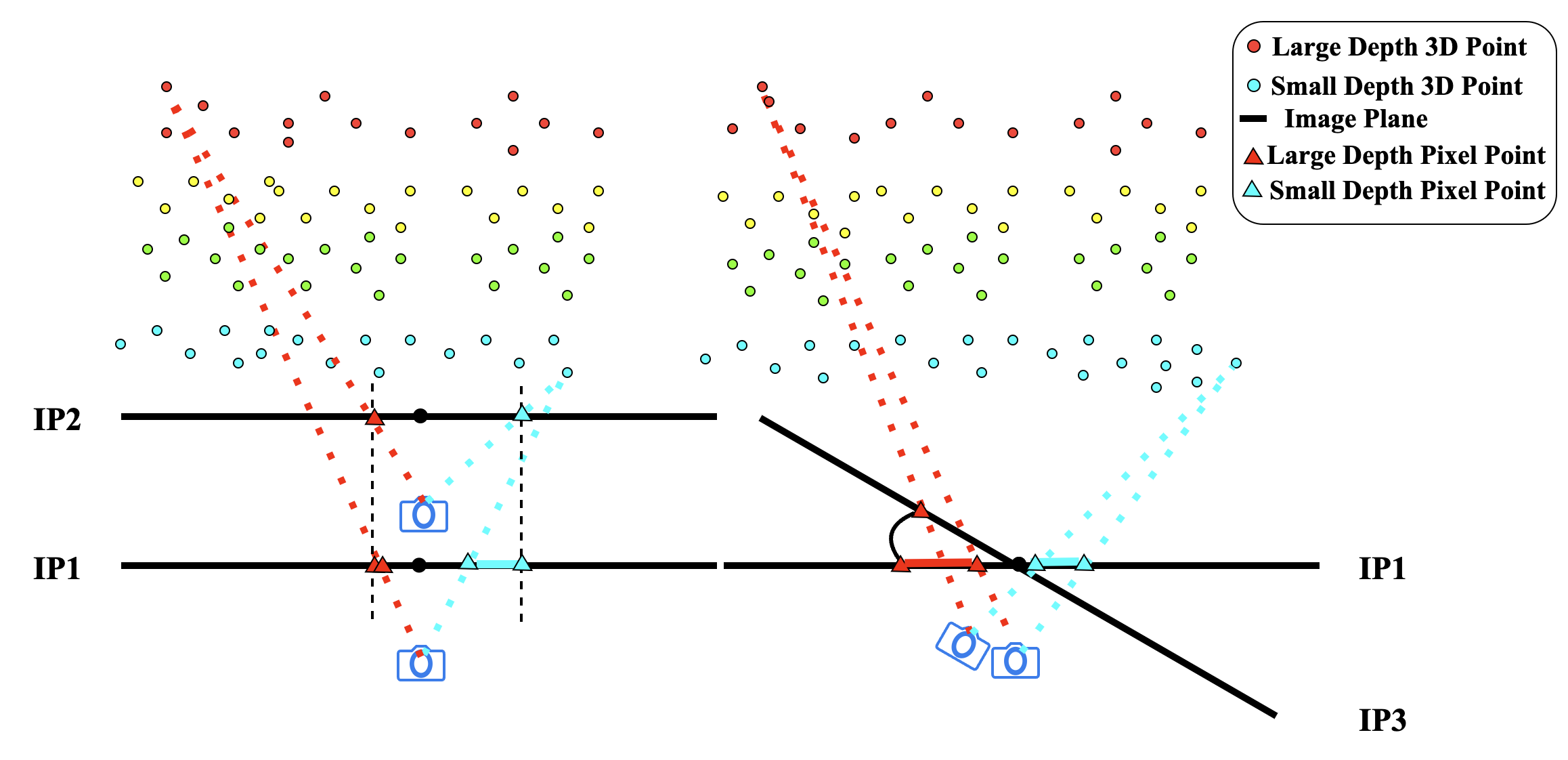}
    \caption{Sensitivity differences of camera rotation and translation to landmarks
at varying depths. The centers of “IP1” and “IP3” coincide.}
    \label{depth distribution}
\end{figure}

From the partial derivatives in the above equation, we observe that the Jacobian matrix associated with rotational components grows as the feature depth increases, highlighting greater sensitivity to rotation. In contrast, when the depth is shallow, the Jacobian matrix for translation becomes dominant, indicating higher sensitivity to translational motion. This relationship is depicted in Fig. \ref{depth distribution}. To leverage the visual feature observations, the depth range (1–200 meters) is divided into 10 uniform intervals, each covering a 20-meter segment. Feature points are evenly distributed across these intervals, with additional points extracted to maintain an approximate mean count per interval if the features are sparse. To handle occlusion effects or changes in scene depth, the interval size and maximum depth limit are dynamically adjusted based on the currently observed maximum depth of available features. For example, in indoor environments, the maximum depth is typically constrained to within 50 meters, whereas in outdoor environments, it generally exceeds 100 meters. 

The proposed depth association strategy eliminates the need for triangulating visual landmarks, as the depth of feature points is directly obtained from LiDAR point clouds with higher precision. This leads to improved accuracy in pose estimation. Furthermore, ensuring a uniform distribution of feature point depths enables the effective utilization of contributions from features at varying depths to the state estimation process. This approach enhances the robustness and efficiency of the overall pose estimation.

\subsection{Deep Learning-based Visual Frontend}
Current SLAM systems predominantly utilize optical flow for tracking sample points. However, this technique exhibits significant limitations in scenarios involving abrupt or extreme variations in illumination. In our method, the visual frontend leverages deep learning-based algorithms for robust feature extraction and matching, ensuring accurate keypoint correspondence under challenging conditions. Specifically, we integrate SuperPoint for feature detection and description, and LightGlue for efficient and adaptive feature matching. SuperPoint is a self-supervised deep learning model designed to detect and describe keypoints with high reliability\cite{superpoint}. It employs a convolutional neural network (CNN) to extract distinctive and repeatable keypoints along with their corresponding descriptors. Notably, SuperPoint demonstrates strong robustness to varying lighting conditions, as it is trained on large-scale datasets with diverse illumination, scale, and viewpoint changes. This ensures that keypoints are consistently detected across images, even in low-light or overexposed scenarios, making it particularly suitable for real-world applications where lighting may fluctuate. 

\begin{figure}
    \centering
    \includegraphics[width=\linewidth]{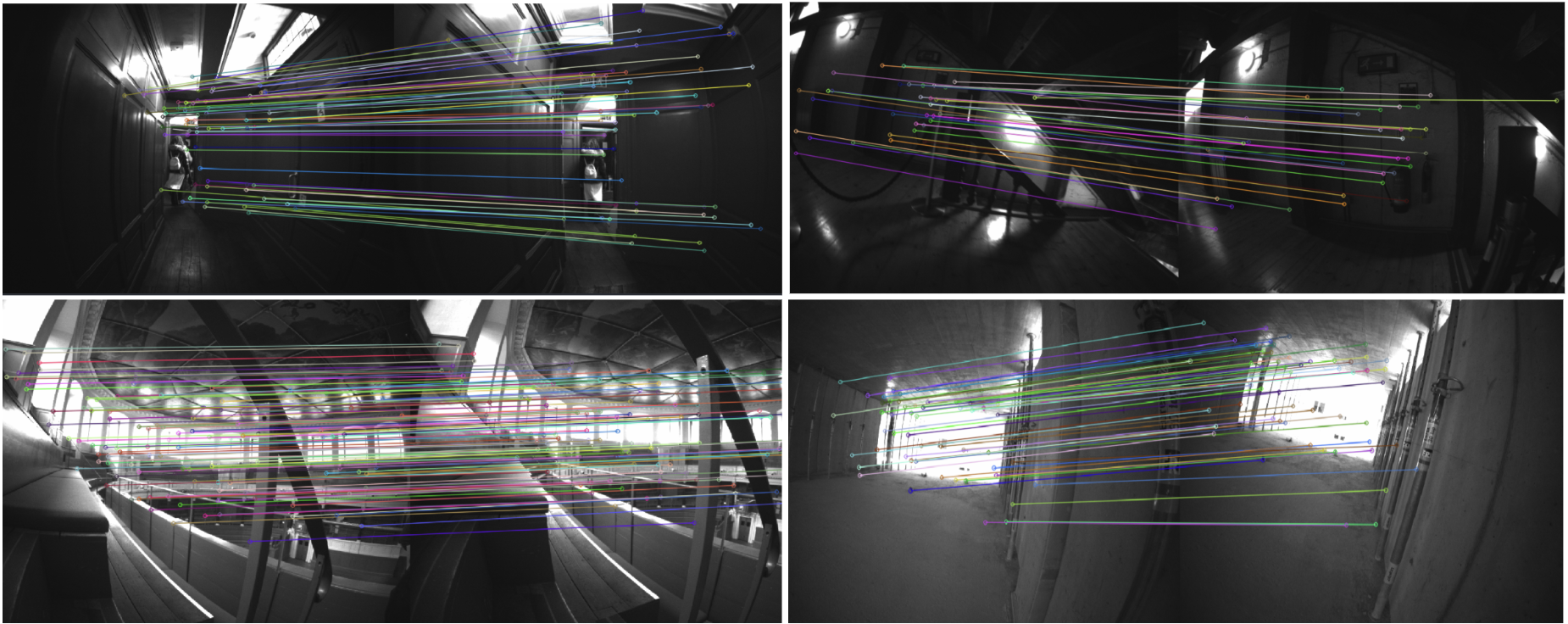}
    \caption{The performance of proposed visual fontend. The images are sourced from Hilti's 2022 attic\_to\_upper\_gallery, construction\_upper\_level, and corridor\_lower\_gallery datasets\cite{hilti}. These datasets were collected in low-light indoor environments and exhibit significant illumination variations.}
    \label{frontend}
\end{figure}

To establish correspondences between SuperPoint keypoints in consecutive frames, we adopt LightGlue, a lightweight transformer-based matching algorithm. LightGlue efficiently prunes irrelevant matches and focuses on high-confidence feature associations through a coarse-to-fine approach. LightGlue is particularly robust under changes in lighting, occlusions, and partial view variations, ensuring stable and reliable feature matching across frames. The performance of our front-end implementation is illustrated in Fig .\ref{frontend}. In low-light indoor environments influenced by point light sources, both optical flow tracking and traditional feature matching algorithms face challenges in achieving reliable and consistent matching results. However, the deep learning-based front-end, due to its robustness to illumination variations, can still produce reliable and consistent matching outputs under challenging lighting conditions\cite{airslam}. Both SuperPoint and LightGlue are deployed using ONNX and NVIDIA TensorRT, utilizing 16-bit floating-point arithmetic for efficient computation.
\subsection{State Update Model}
Similar to FAST-LIO2 and FAST-LIVO, in our method, the state update process primarily utilizes the integration of IMU measurements. The extrinsic parameters and time offsets among the camera, LiDAR, and IMU are pre-calibrated, while the intrinsic parameters of the camera are pre-calibrated, as illustrated in Table \ref{notations}. The IMU frame is designated as the body frame, with the global frame initialized at the position of the first body frame. The discrete state update dynamic model of the system can be described as
\begin{equation}\label{dynamic}
\begin{aligned}
& {x}= \left[\begin{array}{lllllll}
 {R}_i^g &  {p}^g &{v}^g & {b}_g^i & {b}_a^i & {g}^g 
\end{array}\right] \\
& {u}= \left[\begin{array}{ll}
\boldsymbol{\omega}^i & {a}^i
\end{array}\right], {w}= \left[\begin{array}{llll}
{n}_g^i & {n}_a^i & {n}_{{b}_g}^i & {n}_{{b}_a}^i
\end{array}\right] \\
& {f}({x}, {u}, {w})=\left[\begin{array}{c}
\boldsymbol{\omega}^i-{b}_g^i-{n}_g^i \\
{v}^g+\frac{1}{2}\left({R}_i^g\left({a}^i-{b}_a^i-{n}_a^i\right)+{g}^g\right) \Delta t \\
{R_i^g} \left({a}^i-{b}_a^i-{n}_a^i\right)+{g}^g \\
{n}_{{b}_g}^i \\
{n}_{{b}_a}^i \\
{0}_{3 \times 1} 
\end{array}\right]
\end{aligned}
\end{equation}
in which $R_i^g , p^g$ are rotation and translation from IMU frame to global frame; $b_g^i $ and $b_a^i$ denote the IMU gyroscope and accelerator bias in IMU frame, which is modeled as random walk process driven by Gaussian noise $n_{b_g}^i$ and $n_{b_a}^i$, respectively. $ {f}({x}, {u}, {w})$ is the state equation of INS and $\dot{x}=f(x,u,w)$. By integrating (\ref{dynamic}), the state prediction and covariance and be derived, like \cite{fastlio2,fastlivo}. Notably, the parameters we estimate and update is the error state of the $x$, for which we can apply the ESIKF mehtod\cite{fastlio2,R3live,fastlivo}.

\subsection{Measurement Update Model}
The measurement update process is divided into two step: the first step estimates the state using LiDAR point-to-plane residuals, while the second step refines the state using visual reprojection errors.
\subsubsection{LiDAR Measurement Update}: As illustrated in Fig.~\ref{sweeprecombination}, the original LiDAR scan is decomposed and then recombined to achieve synchronization with the camera frame.After synchronization, backward propagation is performed to correct motion distortion, ensuring that the continuously sampled LiDAR points can be treated as being recorded simultaneously with the camera frame. Given a LiDAR point $p^l$, it is first transformed into the IMU frame $p^i$ using the extrinsic calibration between the LiDAR and IMU. Next, with the prior pose provided by the INS, $p^i$ is further transformed into the global frame $p^g$. We assume that $p^g$ lies on a local plane defined by its five nearest neighboring points in the map. Therefore, the point-to-plane residual for $p^g$ is expected to be zero, which can be expressed as:
\begin{equation}
    r_{p^l}(x,p^l) = n^T(R_i^g(R_l^i p^l +t_l^i)+t_i^g - p^g_c)  =0
\end{equation}
where $n$ and $p^g_c$ represent the normal vector and central point of the neighboring plane, respectively. The search for the five closest points is performed using an ikd-tree\cite{fastlio2}. Once the LiDAR measurement update converges, the LiDAR points are transformed into the camera frame using $T_l^c$, in preparation for the subsequent visual measurement update.
\subsubsection{Visual Measurement Update}
\begin{figure}
    \centering
    \includegraphics[width=\linewidth]{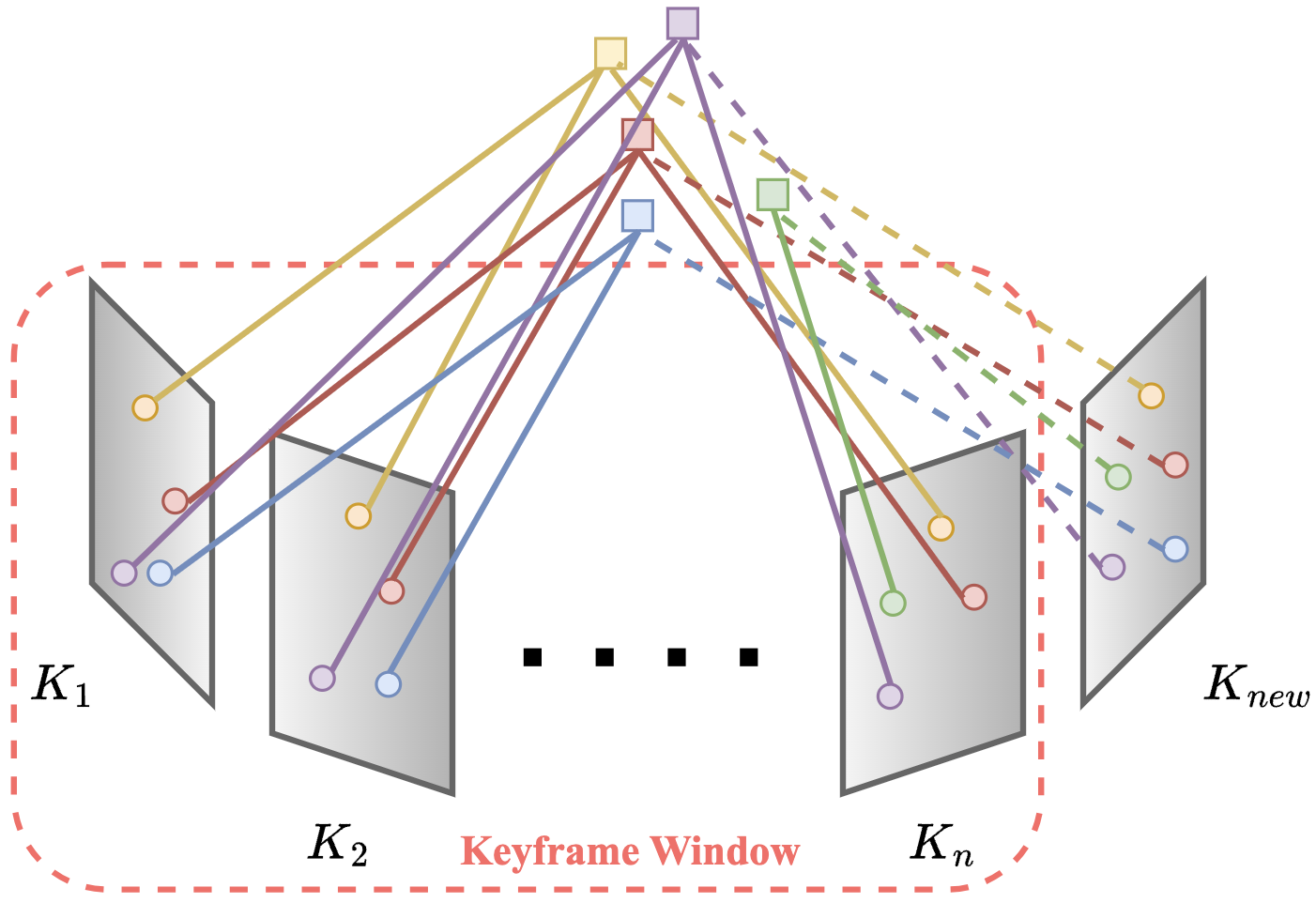}
    \caption{The framework of the visual module. The circles represent feature points extracted by SuperPoint, while the squares denote the corresponding 3D landmarks. The solid lines indicate point pairs with known depths obtained through depth association.}
    \label{framewindow}
\end{figure}

Our visual module employs a sliding window mechanism to maintain a fixed number of keyframes $\{K_1,K_2....K_n\}$, with each keyframe $K_i$ containing pose information, feature points, feature descriptors, and corresponding feature point depths. For a newly added image, its selection as a keyframe is determined based on the variation in its prior pose relative to the $K_1$ keyframe. If the image meets the criteria, feature points are extracted with SuperPoint, and depth association is performed using the LiDAR point cloud captured at the same timestamp. Pose estimation is then performed by matching features with LightGlue and minimizing reprojection residuals between the new keyframe $K_{new}$ and the keyframes $K_i$ within the sliding window. Notably, only keyframes in the sliding window with a parallax larger than 15 pixels are included in the pose estimation process to ensure accuracy and efficiency. The estimatoin strategy employed in the visual module is illustrated in the Fig .\ref{framewindow}.

In our visual measurement update, the construction and optimization of the visual reprojection error are inspired by the approach used in VINS-Mono\cite{vinsmono}. However, with the assistance of depth association from LiDAR point clouds, the feature point depths are accurately and reliably determined. As a consequence, during the optimization process, the poses of $K_i,i=1,2...n$ within the sliding window and the depths of feature points are treated as fixed constants, while only the pose of the newly added $K_{new}$ is optimized. For two consecutive frames $j$ and $k$, the reprojection process maps a feature point $p^{p_j}$ from the pixel plane of frame $j$ onto frame $k$. This serves as the basis for the visual measurement model, which can be formulated as

\begin{equation}
    \begin{cases}
    r_c (p^{p_k},x) = [b1,b2]^T \cdot \left(p^{c_k}- \frac{\tilde{p}^{c_k}}{||\tilde{p}^{c_k}||}\right)\\
    p^{c_k} = \pi_c^{-1}\left(\begin{bmatrix}
        p^{p_k}_u\\p^{p_k}_v
    \end{bmatrix}\right)\\
\begin{aligned}
\tilde{p}^{c_k}={R}_i^c\left( { R } _ { g } ^ { i _ { k } } \left({R}_{i_j}^g( \right.\right. & {R}_c^i \frac{1}{\delta_j} \pi_c^{-1}\left(\left[\begin{array}{l}
 p^{p_j}_u \\
p^{p_j}_v
\end{array}\right]\right) \\
& \left.\left.\left.+{t}_c^i\right)+{t}_{i_j}^g-{t}_{i_k}^g\right)-{t}_c^i\right)
\end{aligned}
    \end{cases}
\end{equation}
in which $\left[p^{p_j}_u,p^{p_j}_v  \right]^T,\left[p^{p_k}_u,p^{p_k}_v  \right]^T$ denote the coordinates in pixel plane of corresponding matched features in frame $j$ and $k$, respectively; $[b_1,b_2]$ are two arbitrarily selected orthogonal bases which span the tangent plane of $\tilde{p}^{c_k}$; $R_{i_j}^g$ and $t_{i_j}^g$ are rotation and translation components from IMU frame to global frame at the time of $K_j$. By treating the poses $R_{i_j}^g,t_{i_j}^g$ of the sliding window keyframes and the feature point inverse depths $\delta_j$ as fixed constants, the number of variables involved in the optimization process is significantly reduced. This not only improves computational efficiency but also enhances the stability and convergence of the optimization process, as it focuses solely on estimating the pose of the newly added keyframe $K_{k}$.
\section{experiments and results}
\subsection{Datasets for Evaluation}
We introduce several datasets utilized for performance evaluation and analysis, including FAST-LIVO-Dataset\cite{fastlivo}, R3LIVE-Dataset\cite{R3live}, NTU-VIRAL\cite{ntu}, as well as Hilti'22\cite{hilti}. The datasets are classified into two categories: the first category comprises datasets with reference ground truth, including NTU-VIRAL and Hilti'22, while the second category includes datasets without reference ground truth but with trajectories exhibiting consistent overlap between their start and end points, such as the FAST-LIVO-Dataset and R3LIVE-Dataset. The NTU-VIRAL dataset was collected using a drone, whereas the remaining three datasets were collected in handheld. Specifically, the NTU-VIRAL dataset incorporates a left grayscale camera, a horizontal 16-channel OS1 gen1 LiDAR, and its internal IMU, with ground truth provided by the Leica Nova MS60 MultiStation. The Hilti'22 dataset employs a left grayscale camera, a Hesai PandarXT-32 LiDAR, and an embedded cellphone-grade IMU, with reference ground truth generated using a novel ground truth collection methodology \cite{hilti}. In contrast, both the R3LIVE-Dataset and FAST-LIVO-Dataset utilize RGB cameras, LiVOX AVIA LiDAR sensors, and the internal IMU, without external ground truth systems. All experiments are conducted on a consumer-grade computer equipped with an Intel Core i7-14700K processor, 32 GB of RAM, and an NVIDIA GeForce RTX 4080 Super GPU.
\begin{table}[!h]
    \centering
    \caption{ RMS (meters) of Absolute Translation errors in NTU-VIRAL and Hilti'22 sequences}
    
    \begin{tabularx}{0.5\textwidth}{llllll}
    \toprule
        Sequence & FAST-LIO2&R3LIVE&FAST-LIVO&SR-LIVO&LIR-LIVO\\ \midrule
        eee\_01&0.255&1.056&0.277&0.216&\textbf{0.164} \\ 
        eee\_02&0.194&---&0.208&0.229& \textbf{0.127}\\ 
        eee\_03&0.246&0.518&0.256&\textbf{0.216}&0.261 \\ 
         nya\_01&0.242&0.252&0.307&0.181& \textbf{0.152} \\ 
         nya\_02&0.225&0.299&0.239&\textbf{0.190}&0.253  \\
         nya\_03&0.177&0.327&0.194&0.203& \textbf{0.161} \\
         sbs\_01&0.254&0.527&0.257&\textbf{0.120}&0.152  \\
         sbs\_02&0.273&0.268&0.276&0.222&\textbf{0.163} \\
         sbs\_03&0.251&0.235&0.257&0.209& \textbf{0.140}\\ \midrule 
         exp06&0.101&0.051&0.098&0.075&\textbf{0.038} \\
         exp14&0.152&0.132&0.111&0.126& \textbf{0.107} \\
         exp16&---&---&---&0.753& \textbf{0.528} \\
         exp18&0.828&---&0.196&0.247&\textbf{0.168} \\
         \bottomrule
    \multicolumn{6}{p{251pt}}{Sequences ``Exp06'', ``Exp14'', ``Exp16'', ``Exp18'' are from Hilti'22 dataset; Others are from NTU-VIRAL dataset. ``---'' denotes the failure in corresponding sequence.}
    \end{tabularx}
    \label{gte}
\end{table}
\subsection{Benchmark Experiments}
In this experiment, extensive evaluations were conducted on 20 sequences from NTU-VIRAL, Hilti'22, R3LIVE-Dataset, and FAST-LIVO-Dataset. In addition to our proposed LIR-LIVO method, several state-of-the-art (SOTA) open-source algorithms were used for comparison, including R3LIVE, a dense direct LiDAR-inertial-visual odometry system; FAST-LIO2, a direct LiDAR-inertial odometry system; and FAST-LIVO, a fast direct LiDAR-inertial-visual odometry system; SR-LIVO, a sweep reconstruction LiDAR-inertial-visual odometry system based on R3LIVE. We directly downloaded the SOTA algorithms from their respective GitHub repositories and utilized their recommended indoor and outdoor LiDAR configurations. For SR-LIVE and FAST-LIVO, we extended the original implementations by incorporating a fisheye camera model and a new LiDAR data format. Additionally, in R3LIVE, we disabled the real-time online estimation of the camera's intrinsic and extrinsic parameters to ensure consistency across all estimation strategies. Beyond these modifications, all parameters such as the number of feature points extracted, the pyramid levels in optical flow tracking, and the covariance of visual observations were kept as the default configurations provided by the authors in their source code.
The results of all algorithms are illustrated in Table \ref{gte} and Table \ref{ete}.
\begin{table}[]
    \centering
    \caption{3D End to End Errors (meters)}
    \begin{tabularx}{0.5\textwidth}{llllll}
    \toprule
        Sequence &R3LIVE&FAST-LIVO&SR-LIVO&LIR-LIVO  \\ \midrule
         hku\_campus\_seq\_00 &0.100&0.029&\textbf{0.020 }&0.029 \\
         hku\_campus\_seq\_02 &0.121&0.115&0.053 & \textbf{0.051}\\
         hku\_park\_00&\textbf{0.078} &0.087&0.120 & 0.111\\
         hku\_park\_01&0.537 &0.596& 0.546&\textbf{0.511} \\
         degenerate\_seq\_00 &0.067&13.003&0.103 &\textbf{0.049} \\
         degenerate\_seq\_01  &0.094&---& 0.091&\textbf{0.084} \\
        LiDAR\_Degenerate &0.064&\textbf{0.044}&0.053 & 0.076\\
        \bottomrule
    \end{tabularx}
    \label{ete}
\end{table}

As shown in Table \ref{gte}, the LIR-LIVO algorithm achieves the lowest RMS (Root Mean Square) absolute translation errors across most sequences, highlighting its superior accuracy compared to other state-of-the-art methods. For the NTU-VIRAL dataset, LIR-LIVO consistently outperforms competitors such as FAST-LIO2, R3LIVE, and FAST-LIVO. For instance, in the ``eee\_01" sequence, LIR-LIVO achieves an error of 0.139 m, significantly lower than FAST-LIO2 and FAST-LIVO. On the Hilti'22 dataset, LIR-LIVO demonstrates exceptional performance, achieving an error of 0.038 m in ``Exp06," significantly better than SR-LIVO and FAST-LIVO. These results highlight the robustness and accuracy of LIR-LIVO in handling both NTU-VIRAL and Hilti'22 sequences, consistently surpassing existing approaches in diverse environments. For the Hilti dataset comparison, R3LIVE was excluded from testing due to the absence of an external IMU configuration. Sequences "Exp14," "Exp16," and "Exp18" were recorded in indoor environments, which present significant challenges for LiDAR-inertial-visual odometry systems. As illustrated in Fig .\ref{frontend}, the Hilti'22 dataset features poor lighting conditions, further increasing the difficulty for visual fontend and pose estimation, which causes both FAST-LIVO and SR-LIVO to perform poorly, leading to failures on the challenging sequences. Table \ref{ete} summarizes the 3D end-to-end errors (meters) across various sequences. SR-LIVO achieves the lowest errors in ``hku\_campus\_seq\_00", while LIR-LIVO shows better performance with 0.051 m in ``hku\_campus\_seq\_02" and excels in ``degenerate\_seq\_00". FAST-LIVO struggles significantly on ``degenerate\_seq\_00". R3LIVE performs well in simpler cases like ``hku\_park\_00" but is less consistent overall. LIR-LIVO demonstrates its robustness in most scenarios, especially in challenging conditions.

\subsection{Time Consumption Analysis}
As shown in Table \ref{time}, LIR-LIVO demonstrates efficient time consumption across datasets, the VIO subsystem contributes the most to computation time, while the LIO subsystem remains efficient. Although the frontend leverages a deep learning-based approach, Table \ref{time} reveals that, with the augmentation of GPU in speed, it is more efficient than the traditional method such as Shi-Tomasi corner and Brute-Force matcher. The visual frontend contributes nearly two-third of the total computation time within the VIO subsystem. The majority of the time is consumed by the feature matching between frames in sliding window and new frame, which 
is primarily determined by the size of the sliding window. The time consumptions in Table \ref{time} are measured with a sliding window size of 5.
\begin{table}[]
    \centering
    \caption{Time consumption (milliseconds) in Each dataset of LIR-LIVO}
    \label{time}
    \begin{tabularx}{0.5\textwidth}{lllll}
    \toprule
        Component & NTU\_VIRAL & Hilti'22& R3LIVE-Dataset \\ \midrule
        SuperPoint & 2.97& 2.71&3.20\\
        Shi-Tomasi & 3.23& 3.12& 3.57\\
        Depth Association & 1.56&1.03&1.69\\
        LightGlue &7.01&9.09&10.47\\
        BFMatcher &11.43&13.96& 15.01\\
        ESIKF Update &0.10&0.12&0.12\\
        VIO Subsystem&11.64&12.95&15.48\\
        LIO Subsystem&5.70&5.39&4.09\\
        LIVO Total &17.34&18.34& 19.57\\
        \bottomrule
    \end{tabularx}
\end{table}

\begin{table}[!h]
    \centering
    \label{ablation}
    \caption{RMS of Absolute Translation errors in NTU-VIRAL dataset with/without Uniform depth distribution}
    \begin{tabularx}{0.5\textwidth}{cccc}
    \toprule
        Method & eee01/02/03&nya01/02/03&sbs01/02/03 \\ \midrule
         LIR-LIVO(w/o)&0.164/0.131/\textbf{0.261}&0.152/0.253/0.210&0.153/0.163/\textbf{0.140}\\
         LIR-LIVO&\textbf{0.151}/\textbf{0.129}/0.266&\textbf{0.148}/\textbf{0.241}/\textbf{0.206}&\textbf{0.152}/\textbf{0.158}/0.142\\
         \bottomrule
    \end{tabularx}
\end{table}
\subsection{Depth Distribution Influence}
Ablation studies were conducted on the NTU-VIRAL dataset to evaluate the impact of enforcing a uniform depth distribution among visual feature points. Specifically, depth-associated feature points within each frame were sorted and uniformly downsampled to 50 points. This strategy significantly reduced the dimensionality of the matrices involved in ESIKF updates, thereby improving computational efficiency. Experimental results, as summarized in the Table \ref{ablation}, indicate that this method outperforms the original approach, with improvements on seven sequences.

\section{conclusion}
This work introduces LIR-LIVO, a novel LiDAR-inertial-visual odometry system designed to address challenges in environments with degraded LiDAR signals and complex illumination conditions. The proposed system combines LiDAR depth association with uniform depth distribution, deep learning based visual feature, and a lightweight visual subsystem to achieve efficient and robust state estimation. Comprehensive evaluations on benchmark datasets demonstrate that LIR-LIVO outperforms in LiDAR degenerated environment. Future work will explore integrating additional sensor modalities and optimizing real-time performance for resource-constrained platforms.


\begin{thebibliography}{1}
\bibliographystyle{IEEEtran}

\bibitem{Unified}
Wisth, David, et al. "Unified multi-modal landmark tracking for tightly coupled LiDAR-inertial-visual odometry." IEEE Robotics and Automation Letters 6.2 (2021): 1004-1011.
\bibitem{LOAM}
Zhang, Ji, and Sanjiv Singh. "LOAM: LiDAR odometry and mapping in real-time." Robotics: Science and systems. Vol. 2. No. 9. 2014.
\bibitem{fastlio2}
Xu, Wei, et al. "Fast-lio2: Fast direct LiDAR-inertial odometry." IEEE Transactions on Robotics 38.4 (2022): 2053-2073.
\bibitem{liosam}
Shan, Tixiao, et al. "Lio-sam: Tightly-coupled LiDAR inertial odometry via smoothing and mapping." 2020 IEEE/RSJ international conference on intelligent robots and systems (IROS). IEEE, 2020.

\bibitem{vinsmono}
Qin, Tong, Peiliang Li, and Shaojie Shen. "Vins-mono: A robust and versatile monocular visual-inertial state estimator." IEEE transactions on robotics 34.4 (2018): 1004-1020.

\bibitem{R3live}
Lin, Jiarong, and Fu Zhang. "R3LIVE: A Robust, Real-time, RGB-colored, LiDAR-Inertial-Visual tightly-coupled state Estimation and mapping package." 2022 International Conference on Robotics and Automation (ICRA). IEEE, 2022.
\bibitem{levins}
Tang, Hailiang, et al. "LE-VINS: A robust solid-state-LiDAR-enhanced visual-inertial navigation system for low-speed robots." IEEE Transactions on Instrumentation and Measurement 72 (2023): 1-13.
\bibitem{srlivo}
Yuan, Zikang, et al. "SR-LIVO: LiDAR-Inertial-Visual Odometry and Mapping With Sweep Reconstruction." IEEE Robotics and Automation Letters (2024).

\bibitem{camvox}Zhu, Yuewen, et al. "Camvox: A low-cost and accurate lidar-assisted visual slam system." 2021 IEEE International Conference on Robotics and Automation (ICRA). IEEE, 2021.

\bibitem{lvisam}
Shan, Tixiao, et al. "Lvi-sam: Tightly-coupled LiDAR-visual-inertial odometry via smoothing and mapping." 2021 IEEE international conference on robotics and automation (ICRA). IEEE, 2021.
\bibitem{fastlivo2}
Zheng, Chunran, et al. "Fast-livo2: Fast, direct LiDAR-inertial-visual odometry." IEEE Transactions on Robotics (2024).
\bibitem{fastlivo}
Zheng, Chunran, et al. "Fast-livo: Fast and tightly-coupled sparse-direct LiDAR-inertial-visual odometry." 2022 IEEE/RSJ international conference on intelligent robots and systems (IROS). IEEE, 2022.

\bibitem{Legoloam}
Shan, Tixiao, and Brendan Englot. "Lego-loam: Lightweight and ground-optimized LiDAR odometry and mapping on variable terrain." 2018 IEEE/RSJ International Conference on Intelligent Robots and Systems (IROS). IEEE, 2018.

\bibitem{stereoLiDAR}
Shao, Weizhao, et al. "Stereo visual inertial LiDAR simultaneous localization and mapping." 2019 IEEE/RSJ international conference on intelligent robots and systems (IROS). IEEE, 2019.
\bibitem{licfusion2}
Zuo, Xingxing, et al. "Lic-fusion 2.0: LiDAR-inertial-camera odometry with sliding-window plane-feature tracking." 2020 IEEE/RSJ International Conference on Intelligent Robots and Systems (IROS). IEEE, 2020.
\bibitem{Droidslam}
Teed, Zachary, and Jia Deng. "Droid-slam: Deep visual slam for monocular, stereo, and rgb-d cameras." Advances in neural information processing systems 34 (2021): 16558-16569.
\bibitem{supslam}
Quach, Cong Hoang, et al. "SupSLAM: A robust visual inertial SLAM system using SuperPoint for unmanned aerial vehicles." 2021 8th NAFOSTED Conference on Information and Computer Science (NICS). IEEE, 2021.

\bibitem{superpoint}
DeTone, Daniel, Tomasz Malisiewicz, and Andrew Rabinovich. "Superpoint: Self-supervised interest point detection and description." Proceedings of the IEEE conference on computer vision and pattern recognition workshops. 2018.
\bibitem{airslam}
Xu, Kuan, et al. "Airslam: An efficient and illumination-robust point-line visual slam system." arXiv preprint arXiv:2408.03520 (2024).
\bibitem{superglue}
Sarlin, Paul-Edouard, et al. "Superglue: Learning feature matching with graph neural networks." Proceedings of the IEEE/CVF conference on computer vision and pattern recognition. 2020.
\bibitem{lightglue}
Lindenberger, Philipp, Paul-Edouard Sarlin, and Marc Pollefeys. "Lightglue: Local feature matching at light speed." Proceedings of the IEEE/CVF International Conference on Computer Vision. 2023.

\bibitem{Xfeat}
Potje, Guilherme, et al. "XFeat: Accelerated Features for Lightweight Image Matching." Proceedings of the IEEE/CVF Conference on Computer Vision and Pattern Recognition. 2024.
















\bibitem{depth enhance}
Zhang, Ji, Michael Kaess, and Sanjiv Singh. "A real-time method for depth enhanced visual odometry." Autonomous Robots 41 (2017): 31-43.

\bibitem{high robust low drift}
Zhang, Ji, and Sanjiv Singh. "Laser–visual–inertial odometry and mapping with high robustness and low drift." Journal of field robotics 35.8 (2018): 1242-1264.

\bibitem{R2live}
Lin, Jiarong, et al. "R2LIVE: A Robust, Real-Time, LiDAR-Inertial-Visual Tightly-Coupled State Estimator and Mapping." IEEE Robotics and Automation Letters 6.4 (2021): 7469-7476.






\bibitem{hilti} Zhang, Lintong, et al. "Hilti-oxford dataset: A millimeter-accurate benchmark for simultaneous localization and mapping." IEEE Robotics and Automation Letters 8.1 (2022): 408-415.


\bibitem{ntu} Nguyen, Thien-Minh, et al. "Ntu viral: A visual-inertial-ranging-LiDAR dataset, from an aerial vehicle viewpoint." The International Journal of Robotics Research 41.3 (2022): 270-280.

\end{thebibliography}
\end{document}